\RequirePackage[hyphens]{url}
\documentclass[letterpaper, 10 pt, conference]{ieeeconf}
\IEEEoverridecommandlockouts %
\makeatletter
\let\MYcaption\@makecaption
\makeatother
\usepackage[font=footnotesize]{subcaption}
\makeatletter
\let\@makecaption\MYcaption
\makeatother

\usepackage{epsfig}
\usepackage{endnotes}

\usepackage[utf8]{inputenc}
\usepackage[utf8]{luainputenc}

\usepackage{ifthen}

\usepackage[table]{xcolor}
\usepackage{diagbox}

\usepackage{nicefrac}

\newif\ifcomments
\commentsfalse
\ifcomments
\usepackage[textsize=small,color=lightgray]{todonotes}
\else
\usepackage[disable]{todonotes}
\fi

\usepackage{textcomp}

\usepackage{pgfplots}
\pgfplotsset{compat=1.13}

\let\labelindent\relax
\newcommand{\subparagraph}{}
\usepackage{enumitem}

\usepackage{etoolbox}
\makeatletter
\patchcmd{\ttlh@hang}{\parindent\z@}{\parindent\z@\leavevmode}{}{}
\patchcmd{\ttlh@hang}{\noindent}{}{}{}
\makeatother

\usepackage{microtype}

\usepackage{balance}

\usepackage{datetime}

\usepackage{listings}

\usepackage{tikz}
\usetikzlibrary{calc}
\usetikzlibrary{trees}
\usetikzlibrary{positioning}
\usetikzlibrary{arrows}
\usetikzlibrary{arrows.spaced}
\usetikzlibrary{arrows.meta}
\usetikzlibrary{chains, matrix}
\usetikzlibrary{shapes, shapes.geometric, shapes.symbols, shapes.multipart}
\usetikzlibrary{decorations.markings, decorations.pathreplacing, decorations.pathmorphing, decorations.text, decorations.shapes}
\usetikzlibrary{backgrounds}
\usetikzlibrary{patterns}
\usetikzlibrary{fit}

\usepackage{amsfonts}
\usepackage{amsmath}
\usepackage{amssymb}
\usepackage{pifont}

\usepackage{mathptmx}

\usepackage{xfrac}

\usepackage[hyphens]{url}

\usepackage{cite}

\usepackage{rotating}
\usepackage{booktabs}
\usepackage{xparse}

\usepackage{outlines}

\usepackage{xspace}
\usepackage{framed}

\usepackage{setspace}

\usepackage{adjustbox}

\newboolean{draft}
\setboolean{draft}{false}

\ifcomments
\paperwidth=\dimexpr \paperwidth + 5cm\relax
\oddsidemargin=\dimexpr\oddsidemargin + 2.5cm\relax
\evensidemargin=\dimexpr\evensidemargin + 2.5cm\relax
\marginparwidth=\dimexpr \marginparwidth + 2.5cm\relax
\fi

\newcommand{\myparagraph}[1]{\noindent{\bfseries #1.}}

\NewDocumentCommand{\rot}{O{45} O{1em} m}{\makebox[#2][l]{\rotatebox{#1}{#3}}}

\usepackage{comment}

\usepackage{graphicx,color}

\DeclareMathAlphabet{\mathcal}{OMS}{cmsy}{m}{n}

\newcommand{\one}{({\em i}\/)}
\newcommand{\two}{({\em ii}\/)}
\newcommand{\three}{({\em iii}\/)}
\newcommand{\four}{({\em iv}\/)}

\newcommand{\pipeline}{Pylot}
\newcommand{\av}{AV}
\newcommand{\carla}{CARLA}

\newcommand{\ap}{AP\textsuperscript{50}}

\usepackage{multirow}

{\end{list}}

\newenvironment{tightenumerate}{%
\begin{list}{\labelenumi}{%
\usecounter{enumi}%
\setlength{\itemsep}{1.5pt}%
\setlength{\topsep}{2pt}%
\setlength{\parskip}{0pt}%
\setlength{\parsep}{0pt}%

\setlength{\labelwidth}{0pt}%
\setlength{\leftmargin}{4pt}%
\setlength{\labelsep}{0pt}%
\setlength{\listparindent}{0pt}%
}}%
{\end{list}}

\usepackage{hyperref}
\hypersetup{
    colorlinks,
    linkcolor={red!50!black},
    citecolor={blue!50!black},
    urlcolor={blue!80!black}
}

\usepackage[capitalize]{cleveref}
\crefformat{section}{#2\S#1#3}
\crefformat{subsection}{#2\S#1#3}
\crefname{section}{\S}{\S}

\title{\pipeline{}: A Modular Platform for Exploring Latency-Accuracy Tradeoffs in Autonomous Vehicles}

\author{\thanks{The authors are affiliated with UC Berkeley.}%
    Ionel Gog\textsuperscript{\textsection}
    \and%
    Sukrit Kalra\textsuperscript{\textsection}
    \and
    Peter Schafhalter\textsuperscript{\textsection}
    \and
    Matthew A. Wright
    \and
    Joseph E. Gonzalez
    \and
    Ion Stoica}
\begin{document}

\date{}
\maketitle
\begingroup\renewcommand\thefootnote{\textsection}
\footnotetext{The three authors have made equal contributions.}
\endgroup
\thispagestyle{empty}
\pagestyle{empty}

\begin{abstract}
    We present \pipeline{}, a platform for autonomous vehicle
    (\av{}) research and development, built with the goal to allow researchers to
    study the effects of the latency and accuracy of their models and algorithms
    on the end-to-end driving behavior of an \av{}.
    This is achieved through a modular structure enabled by our high-performance
    dataflow system that represents \av{} software pipeline components (object
    detectors, motion planners, etc.) as a dataflow graph of operators which
    communicate on data streams using timestamped messages.
    \pipeline{} readily interfaces with popular \av{} simulators like \carla{},
    and is easily deployable to real-world vehicles with minimal code
    changes.

    To reduce the burden of developing an entire pipeline for evaluating a
    single component, \pipeline{} provides several state-of-the-art reference
    implementations for the various components of an \av{} pipeline.
    Using these reference implementations, a \pipeline{}-based \av{} pipeline
    is able to drive a real vehicle, and attains a high score on the
    \carla{} Autonomous Driving Challenge.
    We also present several case studies enabled by \pipeline{}, including 
    evidence of a need for context-dependent components, and per-component time 
    allocation.
    \pipeline{} is open source, with the code available at
    \url{https://github.com/erdos-project/pylot}.

\end{abstract}

\section{Introduction}
\label{s:introduction}
Autonomous vehicles (\av{}s) have attracted considerable research interest and
investment in recent years.
Over the past decade, advances in application areas such as object detection and
localization~\cite{ren2015faster,ssd,tan20efficientdet}, vehicle motion
planning and control~\cite{rrt-star,hybrid-astar,frenet-v2,
  schwarting_planning_2018,guanetti_control_2018}, and behavior and motion
prediction~\cite{lee2017desire,rhinehart2018r2p2,mozaffari_deep_2020}
have enabled rapid advances in \av{} technology.
Reciprocally, the promise of \av{} technology has motivated many recent advances
in emerging technologies like artificial intelligence, with, e.g. driving-based
datasets becoming fundamental baselines for modern computer
vision~\cite{janai_computer_2020,geiger_vision_2013,cityscapes,
  waymo-open-dataset,argo-data}.

\begin{figure}[htb]
    \centering
    \includegraphics[width=\columnwidth]{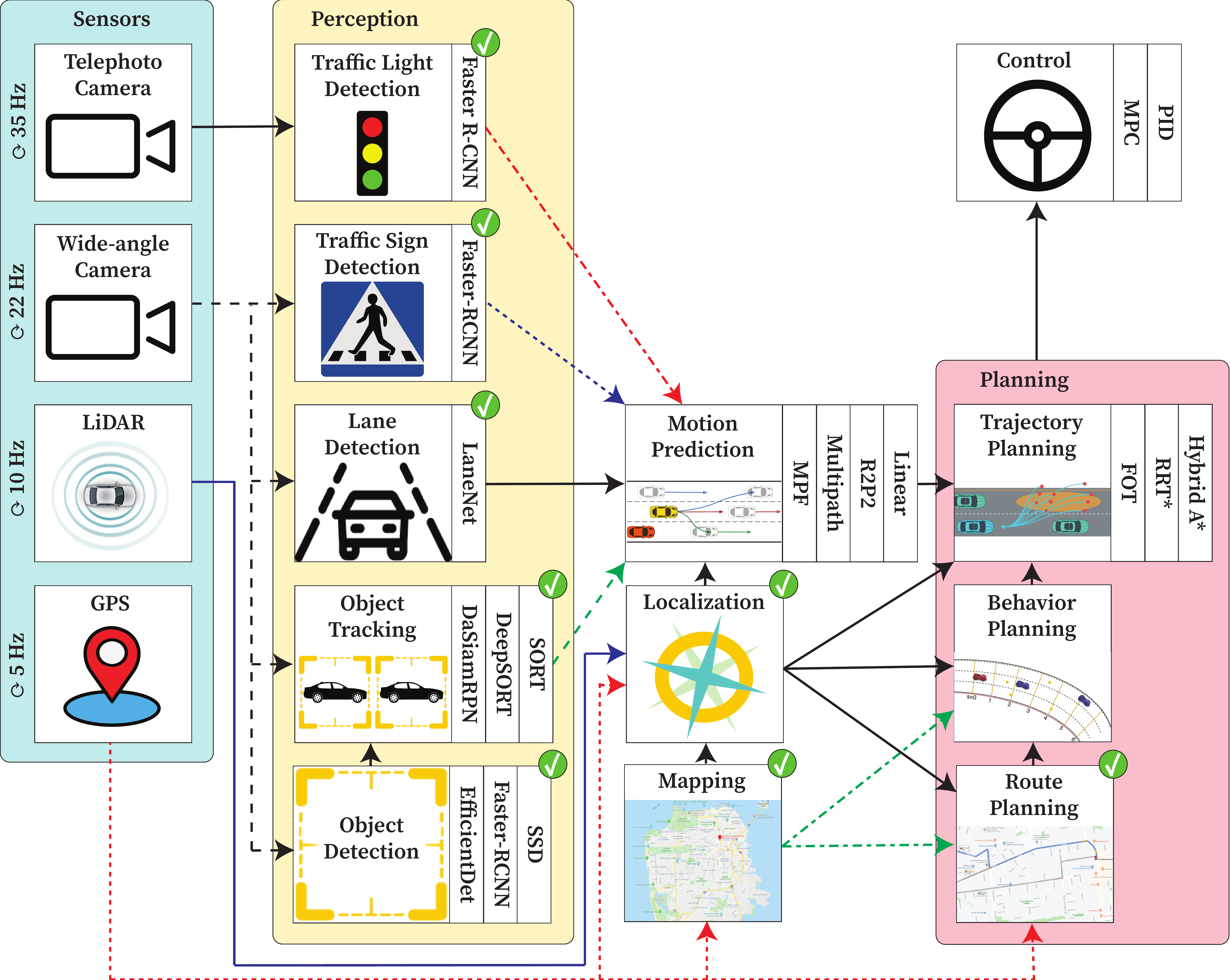}
    \caption{An \av{} pipeline consists of several interconnected modules
	(e.g. perception, planning).  For each component in these modules, 
	\pipeline{} provides reference implementations along with ``perfect'' 
	implementations (for those with a green check mark) that access ground 
	truth data from the simulator.
    }
    \label{fig:schematic_pipeline}
	\vspace{-2.1em}
\end{figure}

Various vendors~\cite{cruise-report,ford-safety-report,ntsb-uber,autoware,
  apollo-baidu} have chosen to realize the computation that underlies an \av{}
as a pipeline similar to the one shown in~\cref{fig:schematic_pipeline}.
In such a pipeline, the output from sensors such as cameras and LiDARs
operating at different frequencies, is used by the perception module to
construct a representation of the environment around the \av{}.
Further, the prediction and planning modules utilize this representation to
construct a trajectory for the \av{} to reach its intended destination while
preventing collisions with other agents and maximizing the comfort of the
\av{}'s passengers.
This modular approach has been the standard for \av{} systems since at
least the DARPA Grand Challenge of the mid-2000s~\cite{thrun2006stanley}.

While this decomposition of the driving task into modules (with multiple
components) has enabled researchers to innovate independently on each module, 
it has also led to the development of problem-specific evaluation metrics that 
fail to account for the end-to-end driving behavior of the 
\av{}~\cite{philion2020learning}.
For example, even driving-based datasets like KITTI~\cite{geiger_vision_2013}
and Cityscapes~\cite{cityscapes}, which are used to develop machine learning 
(ML) models for the perception module, utilize static evaluation metrics such as
average precision~\cite{kitti-detection} that fail to account for the runtime
of the model.
Exploring the tradeoff between the latency of a module and its accuracy on
offline datasets is paramount for safety-critical applications such as
\av{}s where correctness is defined as a function of both the
accuracy of the algorithms and their end-to-end
runtime~\cite{streaming-perception}.

In this paper we introduce \pipeline{}, a modular platform that enables the
study of this critical tradeoff between the accuracy of a module and the
effects of its runtime on the safety of the \av{}.
\pipeline{} is built on top of our high-performance, deterministic
dataflow system~\cite{erdos-code}, and provides:
\one{} state-of-the-art reference implementations for components allowing
researchers to evaluate their algorithms or models in the context of a
realistic \av{} pipeline,
\two{} ``ground truth'' implementations that allow the development of components
and their debugging in the context of an idealized \av{} pipeline, and
\three{} a portable interface that enables seamless transition between a
simulator and a real vehicle.
Finally, \pipeline{} is open source, and is the top submission on
the \carla{} Autonomous Driving Challenge HD map track~\footnote{
  A demo video of \pipeline{} running on the \carla{} Challenge is available
  at \url{https://tinyurl.com/y6ozzpwd}.\label{demo-carla}
}~\cite{carla-challenge}, thus granting the \av{} community a platform
comparable to proprietary pipelines.

The remainder of this paper is organized as follows:
\cref{s:related} gives an overview of existing open AV platforms,
\cref{s:design-criteria} presents the critical design goals of \pipeline{}, and
\cref{s:implementation} discusses how our implementation %
achieves these goals.
Further, \cref{s:pipeline} describes \one{} the reference component 
implementations, \two{} a prototype \av{} pipeline built with these 
implementations, and \three{} our experience of porting this pipeline 
to a real-world \av{}.
Finally, \cref{s:evaluation} presents several case studies %
enabled by \pipeline{} that evaluate in-context performance metrics
of \av{} pipeline components, and 
\cref{s:conclusions} concludes and discusses future work.

\section{Related work}
\label{s:related}
Recent work in object detection has emphasized the need to achieve a balance
between the latency and accuracy of an ML model for a given 
application~\cite{speed-accuracy-detectors}.
While there have been efforts to both define evaluation metrics that integrate
latency and accuracy of the perception module in the context of
\av{}s~\cite{streaming-perception,philion2020learning}, and develop flexible 
backbones for object detection models that enable developers to choose an 
optimum point in the latency-accuracy curve~\cite{tan20efficientdet,yolov4}, 
they fail to account for the effect of the runtime on the end-to-end driving 
behavior.
Moreover, to the best of our knowledge, such metrics do not exist for other 
modules in the \av{} pipeline.

On the other hand, open-source implementations of \av{} pipelines lack 
significantly in their ability to allow developers to explore the accuracy 
of individual components in the context of an end-to-end pipeline.
For example, both Autoware~\cite{autoware} and Baidu's 
Apollo~\cite{apollo-baidu} provide limited interfaces to freely-available
simulation platforms.
Specifically, neither of these \av{} pipelines allow photo-realistic simulation 
of cameras, thus failing to account for the accuracy and runtime of the
perception module.
Moreover, they omit pre-trained models for other components specific to any 
simulation platform which raises the bar for testing a component.
Finally, the debugging ability of these platforms is also limited by their 
choice of the underlying publisher-subscriber communication paradigm which 
complicates the deterministic replay of driving scenarios~\cite{cruise-roscon}.

\section{\pipeline{}'s Design Goals}
\label{s:design-criteria}
The central design goal of \pipeline{} is to support the study of the tradeoff 
between the runtime of components and their accuracy in the context of the
end-to-end driving behavior.
To achieve this goal, \pipeline{} must fulfill three key requirements:

\myparagraph{Modularity}
To enable rich evaluation of new models and algorithms, \pipeline{} must
provide modular components that can be swapped out for alternate
implementations.
This allows developers to compare their components with state-of-the-art
implementations on similar driving scenarios in addition to evaluating their
components using standard metrics on offline datasets.
Moreover, a ``plug-and-play" architecture supports the future introduction
of new modules and components without requiring a tedious overhaul of the
entire pipeline.

\myparagraph{Portability} Developers must be able to transition between 
different simulators and real-world vehicles in order to evaluate their 
components across various driving environments.
For example, a developer should be able to ensure that their planning algorithm
provides similar behavior on a traffic simulator such as
SUMO~\cite{krajzewicz2012recent}, a dynamic world simulator such as
\carla{}~\cite{carla} or AirSim~\cite{airsim2017fsr}, and real-world vehicles.
A key stipulation of portability is that \pipeline{} must be highly performant.
This allows components developed in \pipeline{} to be tested in simulation, and
effortlessly deployed to a real vehicle without any additional changes that 
might affect the tradeoff between latency and accuracy.
On the other hand, the runtimes of components observed in a real vehicle can be 
faithfully reproduced in order to ensure that the latency-accuracy tradeoff can 
be correctly explored in simulation.

\myparagraph{Debuggability} Ensuring the safety of models and algorithms
requires extensive testing across various scenarios.
Thus, \pipeline{} must provide developers with tools that allow them
to understand and easily debug their components when they exhibit abnormal
or unsafe behavior.
In order to reproduce unsafe behaviors, the software system must be
output deterministic (i.e. produce the same output given the same
inputs)~\cite{altekar2009odr}, and the pipeline must enable seamless logging
of the data necessary to reconstruct the behavior of the \av{} pipeline.

\section{Achieving Design Goals}
\label{s:implementation}
We realize the design requirements outlined in~\cref{s:design-criteria} in our
implementation of \pipeline{}, which consists of approximately 28,000 lines of
Python code.
While \pipeline{} executes atop our low-overhead open-source streaming dataflow
system implemented in Rust~\cite{erdos-code}, we chose to implement \pipeline{} 
itself in Python in order to enable faster prototyping and easier interfacing to 
both simulators such as \carla{}~\cite{carla} and deep learning frameworks such 
as PyTorch~\cite{pytorch} and Tensorflow~\cite{tensorflow}.
The remainder of this section discusses the design of \pipeline{} by focusing on
how it achieves \emph{modularity} (\cref{ss:modularity}), \emph{portability}
(\cref{ss:portability}) and \emph{debuggability} (\cref{ss:debuggability}).

\subsection{Modularity}
\label{ss:modularity}
\pipeline{}'s modular structure is achieved through a dataflow
programming model, where the \av{} pipeline is structured as a directed
graph in which vertices, also known as \emph{operators}, perform computation
(e.g. running object detection) and edges, also known as \emph{streams},
enable communication through timestamped messages (e.g. transmitting bounding
boxes of detected objects).
This inter-component communication style is reminiscent of the familiar
``publisher-subscriber'' model with the operators akin to ROS nodes,
and streams akin to the ROS publishers and subscribers.
Similar to the publisher-subscriber model, the dataflow programming model
limits interactions between operators to streams thus allowing them to be
swapped as long as they conform to the same interface (e.g. an object
detection component based on Faster-RCNN~\cite{ren2015faster} can be swapped
for one based on EfficientDet~\cite{tan20efficientdet}).

However, unlike the publisher-subscriber model, the dataflow programming model
allows swapping of components that differ in their runtimes and resource
requirements without requiring cascading changes throughout the entire pipeline.
Specifically, while a ROS node synchronizes incoming data from multiple sources
by fetching data from the required publishers at a fixed frequency,
a dataflow system allows developers to register callbacks
that get invoked upon the arrival of synchronized data across the requested
streams.
The dataflow system underneath \pipeline{} seamlessly synchronizes data across
multiple streams by requiring the operators that publish on those streams to
send a special \emph{watermark} message upon completion of the outgoing data
for a specific timestamp~\cite{punctuations,flink,naiad,google-cloud-dataflow}.
Hence, while swapping a component with a different runtime in the
publisher-subscriber model requires fine-tuning the frequency at which the
downstream operators invoke their computation, a dataflow system is robust to
such runtime variabilities and enables highly modular applications.

\subsection{Portability}
\label{ss:portability}
In order to allow pipelines developed in simulation to be ported to
real-world vehicles, \pipeline{} must support high-throughput processing of
the data generated by a vehicle's sensors~\cite{cruise-report,cruise-roscon,
google-sensor-throughput}.
This is enabled through our custom high-throughput and low-latency dataflow
system that outperforms ROS, a commonly used robotics middleware, by 30\% in
terms of communication latency~\cite{erdos-code} while providing a shim layer
that allows \emph{operators} to interface with legacy ROS code.
The underlying system \emph{transparently} schedules the parallel execution of 
these operators across machines according to their resource requirements, and
provides collocated modules with 
zero-copy communication via shared memory queues.
Moreover, this transparent scheduling and communication does not require
any code changes, and coupled with the shim layer for legacy ROS code allows 
seamless and piecemeal porting of \pipeline{} to different hardware platforms.

Conversely, \pipeline{} also enables the transition from real-world vehicles
to simulation by enabling the precise replication of the runtimes of components
in a simulator.
We achieve this through \pipeline{}'s integration with the \carla{} simulator.
\carla{} provides two modes of execution:
\one{} \emph{synchronous}, where the simulator allows the instantaneous
application of control commands by pausing the simulator after sending sensor
inputs to the client, and
\two{} \emph{asynchronous}, where the simulator moves forward and
applies the control command when \pipeline{} finishes execution.
While \carla{}'s asynchronous mode should allow control commands to be precisely
applied when the computational pipeline finishes executing, it lags behind
due to the high rendering cost of the underlying graphics engine.
This results in the imprecise application of control commands from the pipeline
to the simulated vehicle.

To enable a timely application of the control commands, we developed
a \emph{pseudo-asynchronous} execution mode that allows \pipeline{} to maintain 
tight control over the simulation loop.
In this mode, we run \carla{} \emph{synchronously} with a high frequency 
(over $200$Hz), and attach a synchronizer between the control module and the 
simulator.
This synchronizer tracks the runtime of the pipeline for each sensor input, and
buffers the control command until the simulation time is at least the 
sum of the sensor input time and the runtime of the pipeline.  

Both the synchronizer and \pipeline{} can be ported to other
simulators~\cite{airsim2017fsr,rong2020lgsvl} with minimal effort.
Developers are only required to implement drivers to extract sensor data from 
the simulator (e.g., camera frames, LiDAR point clouds), and an operator to 
send control commands to the simulator.

\subsection{Debuggability}
\label{ss:debuggability}
\pipeline{} integrates with \carla{}'s ScenarioRunner~\cite{carla-scenario},
and provides a suite of test scenarios based on the National Highway Traffic
Safety Administration's pre-crash scenarios~\cite{nhtsa-scenarios}.
In contrast to other \av{} pipelines implemented atop a publisher-subscriber
model~\cite{autoware,cruise-roscon,apollo-baidu}, \pipeline{}'s underlying
dataflow programming model enables deterministic replay of such scenarios,
thus allowing easier debugging of components.
This deterministic execution is achieved by performing computation on the
receipt of \emph{watermark} messages~\cite{naiad,flink,google-dataflow}.
This reduces the execution of the entire \av{} pipeline to a Kahn Process
Network~\cite{kahn-process-network}, which guarantees determinism.

To aid in debugging the latency and accuracy of components, \pipeline{}
provides extensive logging of both output data and the runtime of each
component.
Specifically, \pipeline{} can be configured to log sensor data, and outputs
of internal modules such that scenarios can be reproduced and debugged offline.
Moreover, \pipeline{} provides fine-grained logs for the execution time of
each component that can be visualized using trace profiling
tools~\cite{chrome-profiling} and replayed in order to study the
latency-accuracy tradeoffs in deterministic scenarios.

To aid in development, \pipeline{} provides ``perfect'' modules that obtain
ground-truth data from \carla{}, which can be used to test other modules in
isolation (e.g., planning using perfect perception), and to generate new
training data sets.
Moreover, \pipeline{} can be deployed with different sensor setups
(e.g., cameras only, cameras and LiDARs), and can be used to study the
end-to-end performance of each setup, and as well the robustness
of solutions when only partial or noisy sensor data is
available~\cite{philion2020lift}.

\begin{figure}
  \centering
  \includegraphics[width=1.0\columnwidth]{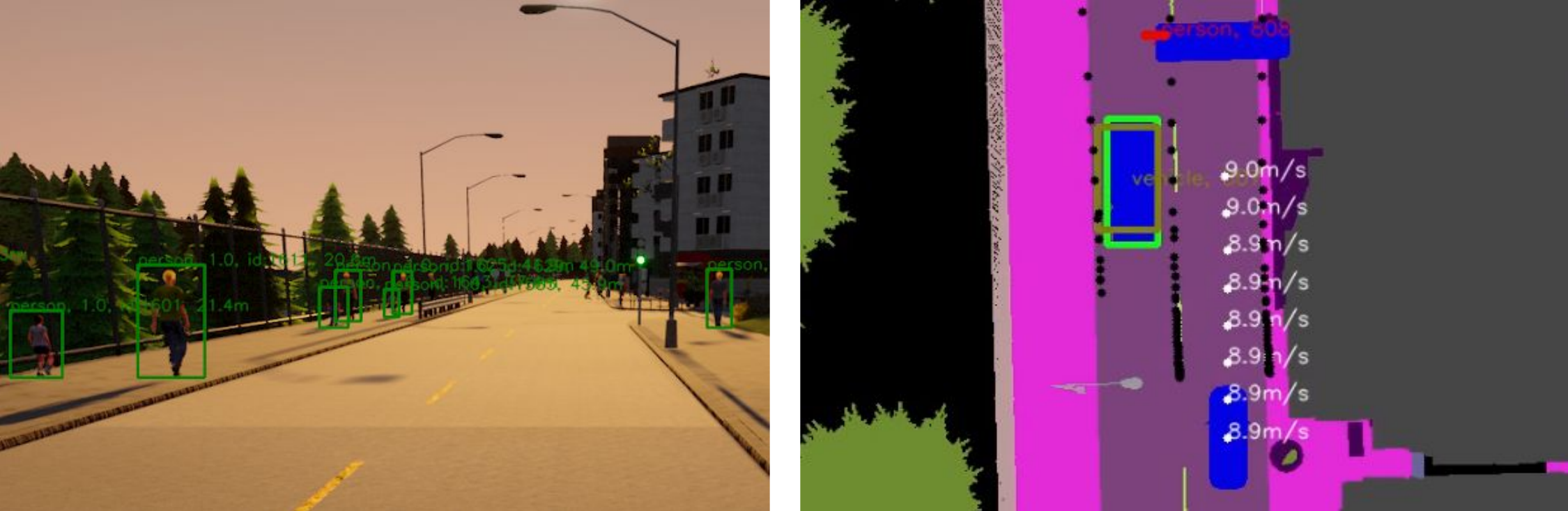}
  \caption{\pipeline{}'s visualizations for critical components.
    The object tracker view (left) shows the bounding boxes and the
    identifiers of detected agents in the camera frame.
    A bird's eye view of the planning module (right) includes lanes,
    predictions for agents, and waypoints computed by the planner.}
  \label{f:planning-world}
  \vspace{-1.75em}
\end{figure}

In addition, \pipeline{} provides visualizations for important
information, such as detected objects, lanes, and traffic lights;
sensors including cameras, LiDARs, and inertial
measurement units; and algorithmic outputs for depth perception,
pose estimation, behavior prediction, semantic segmentation, and planning.
\cref{f:planning-world} exemplifies two such visualizations: \one{}
the output of one of \pipeline{}'s reference object trackers, and \two{}
a bird's eye view of the information used by \pipeline{}'s planners to
compute trajectories (detected obstacles, lanes, traffic signs,
predicted trajectories, and proposed waypoints).

\section{\pipeline: an AV development platform}
\label{s:pipeline}
The goal of \pipeline{} is to accelerate research into new  algorithms and
models for autonomous driving as well as the design of the underlying software
systems.
To support this goal, \pipeline{} provides:
\one{}  state-of-the-art algorithms, pre-trained models, and evaluation metrics,
\two{} the ability to selectively replace components and modules with
ground-truth data from simulators,
\three{} an easily extensible deterministic runtime environment, and
\four{} a range of challenging driving scenarios.
In the rest of this section, we describe the key components and modules in
\pipeline{} (c.f. \cref{fig:schematic_pipeline}), and our experience of porting
\pipeline{} from a simulator to a Lincoln MKZ test vehicle~\footnote{
  A demo video of \pipeline{} running on the MKZ is available
  at \url{https://tinyurl.com/y5vsly3f}.\label{demo}}.

\subsection{Object Detection}
\label{ss:pipeline:perception}
The object detection component comprises of operators that process camera and
LiDAR data to detect, and localize objects, lanes, and traffic lights.
These operators communicate their results via an \texttt{ObstacleMessage} that
contains the bounding box of the detected object along with the confidence
score returned by the ML model.
Users may run multiple versions of these models in parallel in order to
benchmark against each other and against perfect detection by using the
standard accuracy metrics (e.g. mAP) provided by \pipeline{}.

\pipeline{} allows drop-in replacements of models conforming to the Tensorflow
Object Detection API~\cite{speed-accuracy-detectors}, and
provides several configurations of Faster-RCNN~\cite{ren2015faster} and
SSD~\cite{ssd} models trained on data collected from
the \carla{} simulator using \pipeline{}.
In addition, we also utilize the EfficientDet~\cite{tan20efficientdet} family
of models which are built atop of a variable-sized network backbone.
This property of EfficientDet along with the different configurations of
Faster-RCNN and SSD allow the exploration of the
runtime-accuracy tradeoff space.

\subsection{Object Tracking}
\label{ss:pipeline:tracking}
The tracking component estimates the bounding boxes of objects over time,
and maintains consistent identifiers for the objects across detections.
\pipeline{} provides three object trackers that cover the two main
tracking approaches: \one{} \emph{tracking-by-detection} continuously
updates tracker state with bounding boxes received from object detection, and
\two{} \emph{detection-free tracking} follows a fixed number of objects over
time.
These trackers utilize the bounding boxes from the \texttt{ObstacleMessage}
and communicate their results via an \texttt{ObstacleTrajectoryMessage} that
contains the bounding box along with the identifier for each obstacle.

We find that each tracker has context dependent performance and accuracy
tradeoffs.
For example, the SORT~\cite{tracking_sort} tracker is a lightweight multiple
object tracker of the tracking-by-detection variant that uses Kalman filters to
estimate object positions between frames assuming a constant linear velocity
model,
and the Hungarian algorithm to match bounding boxes upon detection updates.
While SORT is fast, it offers low accuracy in the presence of object occlusions
or camera motion.
However, DeepSORT~\cite{tracking_deep_sort} improves on these limitations by
incorporating appearance information for each tracked object, at the cost
of an increased runtime
from executing a convolutional feature extraction model.

On the other hand, DaSiamRPN~\cite{tracking_da_siam_rpn} is a detection-free
single object tracker that leverages a siamese feature extraction network to
learn distractor-aware features.
The tracker relies on neural network inference to track an object thus providing
more accurate estimations between detections at %
the cost of increased runtime.
However, the ``best'' choice of tracker depends on the situation.
For example, SORT performs best in emergency scenarios when low runtime is
critical and DeepSORT is best in regular driving situations, while DaSiamRPN
excels in scenarios that demand high accuracy for a small number of objects
(e.g. urban driving).

\subsection{Prediction}
\label{ss:pipeline:prediction}
The prediction modules uses the tracked history of nearby agents
(vehicles, pedestrians, etc.), along with scene context from LIDAR, to predict
future agent behavior.
Specifically, it receives the bounding box and the identifier of each obstacle
at every time instant through the \texttt{ObstacleTrajectoryMessage}, and
generates an \texttt{ObstaclePredictionMessage} containing the past and the
predicted trajectory for each obstacle.

\pipeline{} contains multiple implementations of the prediction module trained
on \carla{} data that represent the major classes of ML based approaches for
trajectory prediction.
As a baseline, \pipeline{} also provides a linear predictor that utilizes a
linear regression model that assumes that each agent will travel at a constant
velocity and predicts their forward position based on their past trajectory.
While this predictor is fast and hence ideal for emergency collision-avoidance
scenarios, it does not account for the behavior of multiple agents.

Additionally, \pipeline{} provides R2P2~\cite{rhinehart2018r2p2}, a
state-of-the-art single-agent trajectory forecasting model which learns a
distribution over potential future trajectories that is parameterized by a
one-step policy using a gated recurrent unit, and attempts to optimize for both
quality and diversity of samples.
To extend R2P2 to the multi-agent setting, \pipeline{} runs R2P2 on every agent
by rotating the scene context and past trajectories of other agents to the
ego vehicle coordinate frame.

Finally, Multipath~\cite{chai2019multipath} uses a lightweight neural
network to obtain a useful representation of the scene and then applies a
smaller network features for each agent to output predictions.
Because the per-agent computation can be batched, Multipath is fast
but less accurate owing to its inability to explicitly consider agent
interactions.
In contrast, Multiple Futures Prediction (MFP)~\cite{tang2019multiple} jointly
models agent behavior, leading to increased runtime but more accurate
predictions.

\subsection{Planning}
\label{ss:pipeline:planning}
The goal of the planning module is to produce a safe, comfortable, and
feasible trajectory that accounts for the present and future possible states of
the environment.
To achieve this, the planning module in \pipeline{} synchronizes the output
from all the other modules to construct a \texttt{World} representation that
contains the past and future trajectories of all agents in the scene, along with
the location and state of static objects such as traffic lights, traffic signs
etc.
Crucially, this allows \pipeline{} to selectively utilize ground truth
information for any of the previous components or modules and ascertain the
effects of the accuracy and runtime of a selected set of components on the
end-to-end driving behavior of the \av{}.

The planning module is comprised of three components: route, behavioral, and
motion planners, with the latter having the greatest effect on the comfort
and the end-to-end runtime of the \av{}.
Hence, \pipeline{} provides implementations for each of the three main classes
of motion planners:
graph search, incremental search, and trajectory
generation~\cite{lavalle-planning,katrakazas,paden-survey}.

\emph{Graph-based search planners} (e.g. Hybrid A*~\cite{hybrid-astar})
discretize the configuration space as a graph and provide fast results at
lower discretizations. However, a poor choice of the discretization value
may produce infeasible paths.
On the other hand, \emph{incremental search planners} (e.g.
RRT*~\cite{rrt-star}) gradually build a path by sampling the configuration space
instead of precomputing a fixed set of configuration nodes.
They are not limited by the initial graph construction, and thus provide
the ability to fine-tune the accuracy of the result for any given computation
time.
Finally, \emph{trajectory generation planners} (e.g. Frenet Optimal
Trajectory~\cite{frenet-v1,frenet-v2}) construct a set of candidate paths,
which they validate for collisions and physical constraints.
While the resulting paths are usually smoother than their counterparts,
trajectory generation may omit feasible paths if the discretization
is too coarse.

\subsection{Control}
\label{ss:pipeline:control}
The control module receives waypoints and target speeds from the planning
module.
It aims to closely follow the provided waypoints while maintaining its target
speed.
\pipeline{} offers two control options: a Proportional-Integral-Derivative
(PID) controller, and a Model Predictive Control (MPC) controller.
Both options compute commands that adjust brakes, steering, throttle,
and send these to the simulator or a real-world vehicle's drive-by-wire kit.

While \pipeline{} currently comprises of the modules described above, developers
can integrate multi-functionality modules (e.g. MPC for both motion planning
and control~\cite{amos2018differentiable,liu2017path}) or replace entire
subgraphs of the pipeline with end-to-end learning-based
solutions~\cite{liang2020pnpnet, zeng2019end,wayve-website}.

\begin{figure*}
  \begin{subfigure}[b]{0.33\textwidth}
    \centering\captionsetup{width=0.95\linewidth}
    \includegraphics[scale=0.8]{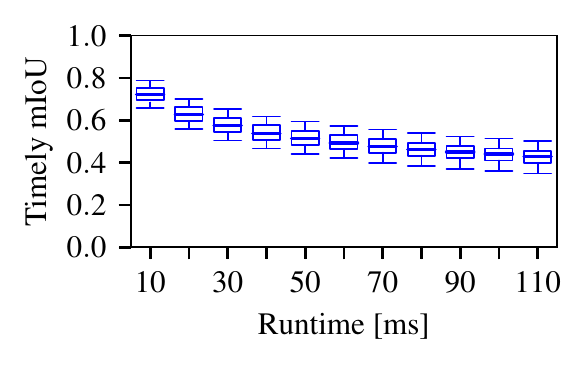}
    \caption{Semantic segmentation timely mIoU.}
    \label{f:real-world-time-mIoU-segmentation}
  \end{subfigure}
  \begin{subfigure}[b]{0.33\textwidth}
    \centering\captionsetup{width=0.95\linewidth}
    \includegraphics[scale=0.8]{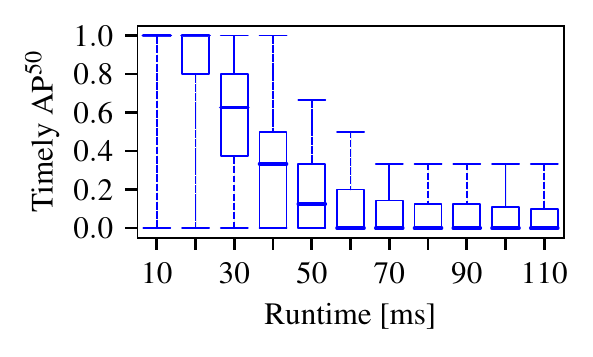}
    \caption{Pedestrian detection timely \ap{}.}
    \label{f:real-world-time-AP50-detector}
  \end{subfigure}
  \begin{subfigure}[b]{0.33\textwidth}
    \centering\captionsetup{width=0.95\linewidth}
    \includegraphics[scale=0.8]{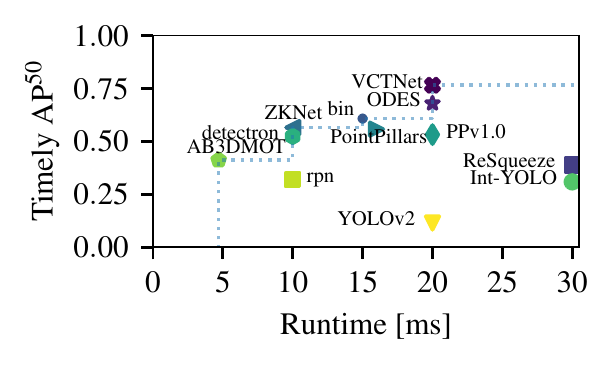}
    \caption{Choice of best detector changes.}
    \label{f:real-world-detector-trade-offs}
  \end{subfigure}
  \begin{subfigure}[b]{0.33\textwidth}
    \centering\captionsetup{width=0.95\linewidth}
    \includegraphics[scale=0.8]{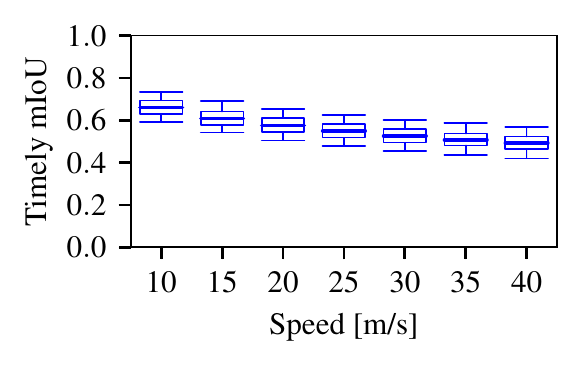}
    \caption{Semantic segmentation timely mIoU degrades faster at
      high speeds.}
    \label{f:real-world-speed-mIoU-segmentation}
  \end{subfigure}
  \begin{subfigure}[b]{0.33\textwidth}
    \centering\captionsetup{width=0.95\linewidth}
    \includegraphics[scale=0.8]{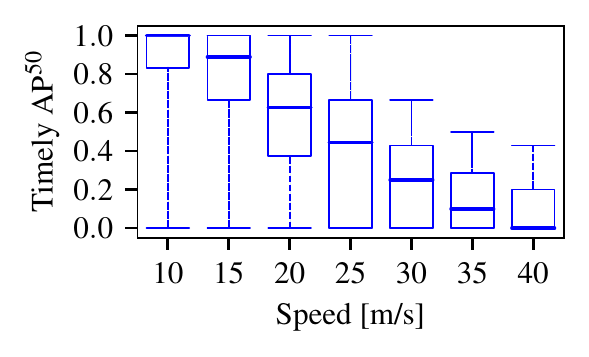}
    \caption{Pedestrian detection timely \ap{} degrades faster at high
      speeds.}
    \label{f:real-world-speed-AP50-detector}
  \end{subfigure}
  \begin{subfigure}[b]{0.33\textwidth}
    \centering\captionsetup{width=0.95\linewidth}
    \includegraphics[scale=0.8]{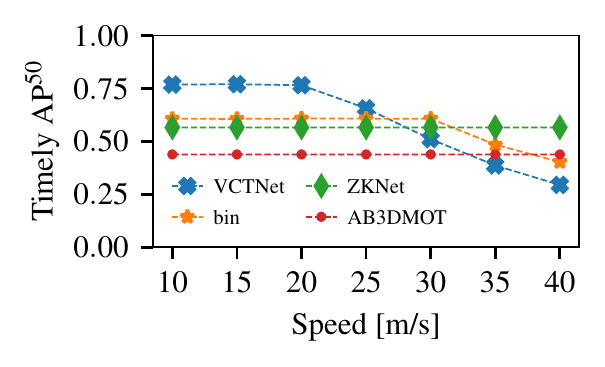}
    \caption{Different object detectors are best at different driving speeds.}
    \label{f:real-world-speed-detector-trade-off}
  \end{subfigure}
  \caption{Timely mIoU and \ap{} degrade when runtime and driving speed
    increase. Thus, accuracy, runtime, and driving speed are all important when
    making model decisions (\ref{f:real-world-detector-trade-offs} and
    \ref{f:real-world-speed-detector-trade-off}).    }
\end{figure*}

\subsection{Deploying \pipeline{} on an Autonomous Vehicle}
\label{ss:pipeline:port}
In order to port \pipeline{} from the \carla{} simulator to a Lincoln MKZ \av{},
we made the following minimal changes:

\begin{tightenumerate}
  \item{} \textbf{Sensors:} we utilized the shim layer that injects data from
  ROS topics exposed by the drivers of the \av{}'s sensors.
  \item{} \textbf{Control:} we modified \pipeline{} to send control commands to
  the ROS node exposed by a drive-by-wire kit~\cite{dataspeed}.
  \item{} \textbf{Localization:} we used the Normal
  Distributions Transform (NDT) algorithm~\cite{biber2003normal}
  implemented in Autoware~\cite{autoware}.
  \item{} \textbf{Model replacement:} we replaced our \carla{}-trained
  detection and tracking models with off-the-shelf models trained on real-world
  datasets.
\end{tightenumerate}

As mentioned previously, porting from simulation to a real vehicle required only
434 lines of Python code.
This efficiency demonstrates the flexibility of our interfaces.

\section{Evaluation}
\label{s:evaluation}
We now show how \pipeline{} enables researchers to study: \one{} the trade-off
between accuracy and runtime for different components
(\cref{ss:eval-accuracy-vs-runtime}), \two{} the effects that
changes in a single component have on an end-to-end driving metric
(\cref{ss:eval-end-to-end}).

\myparagraph{Experimental Setup} The experiments were performed atop
\carla{} on a machine having 2$\times$Xeon Gold 6226 CPUs, $128$GB of DDR4
RAM and 2$\times$Titan-RTX 2080 GPUs.

\subsection{Accuracy \emph{vs.} Runtime}
\label{ss:eval-accuracy-vs-runtime}

We introduce a new family of metrics, \textbf{timely accuracies}, in order to
capture the impact of runtime on accuracy of modules.
The timely accuracy is a metric of how accurate a module's results are with
respect to the present world, and not the past world captured by the input
on which the result was based.
The key idea is to evaluate a module's output performed on inputs at time
$t_1$ with the ground truth labels at $t_2 = t_1+l_t$, where $l_t$ is the
module's runtime.
Thus, timely accuracy captures how accuracy degrades as a function of
runtime in dynamic worlds.

In the experiment, we attached a camera to a simulated \av{}, and set up a
scenario in which the \av{} drove in a city at $20$ m/s.
For each frame we collected the ground-truth semantic segmentation
and computed the \emph{timely} (i.e. time-delayed) mean Intersection over
Union (mIoU) by emulating different prediction runtimes.
\cref{f:real-world-time-mIoU-segmentation} shows that a runtime of $10$ms
is sufficient to reduce mean \emph{timely} mIoU of a perfect semantic
segmentation component to approximately $0.75$, which is less than the mIoUs
of the top three Cityscapes submissions~\cite{cityscapes}.
Similarly, we measured the tradeoff between accuracy and runtime for pedestrian
detectors by computing \emph{timely} Average Precision at IoU 50\% (\ap{}) for a
perfect detection component (i.e. with $1.0$ \ap{}).
\cref{f:real-world-time-AP50-detector} shows that mean \emph{timely} \ap{}
halves with a runtime of $35$ms.
Thus, an object detector with long runtimes must be accompanied by a tracker or
a prediction component that can predict accurate trajectories for longer than
the detector's runtime.

The timely accuracy does not only depend on runtime, but also on the \av{}'s
driving speed.
The faster an \av{} drives, the quicker the world changes.
To show the effect of an \av{}'s speed on timely accuracy, we emulated
a $20$ms runtime for the perfect semantic segmentation and detection components,
and varied the driving speed.
In~\cref{f:real-world-speed-mIoU-segmentation}, we illustrate that
the \emph{timely} mIoU for semantic segmentation decays as speed
increases: median \emph{timely} mIoU is 28\% smaller at $40$m/s than at $10$m/s.
In the case of object detection, speed has a bigger effect on \emph{timely}
\ap{}: median \emph{timely} \ap{} is 1.0 when driving at $10$m/s, but it
decreases to $0$ at $40$m/s (\cref{f:real-world-speed-AP50-detector}).
In~\cref{f:real-world-speed-detector-trade-off} we illustrate the
trade-off between model accuracy and runtime using models from the KITTI
pedestrian detection challenge~\cite{kitti-website}.
We observe that fast, low-accuracy detectors obtain higher timely
accuracy than slow, high-accuracy detectors when driving at high speeds.

\subsection{Effects of Component Changes on End-to-end Driving}
\label{ss:eval-end-to-end}
We now leverage \pipeline{}'s \emph{pseudo-asynchronous} execution mode to study
both the effect of component changes and model hyperparameter tweaks on
end-to-end driving.
For this experiment, we developed a scenario which simulates the illegal
crossing of the \av{}'s lane by a person, as shown in \cref{f:planning-world}.
The scenario is further complicated by the presence of a truck on the
opposite lane, which occludes the person until it reaches the \av{}'s lane
($20$ meters away).
This presents an imminent threat to the \av{}, and requires the \av{} to
perform an emergency maneuver to avoid a collision.

\begin{table}
  \scriptsize
  \caption{Configurations that avoid a collision are marked in
    \textcolor{green}{green}, while failing configurations are marked in
    \textcolor{red}{red}.}
  \centering
  \begin{tabular}{l|c|c|c|c}
  \diagbox{Planner [P\textsubscript{99} runtime]}{Speed [m/s]}  & 16 & 18 & 20 & 22 \\
  \hline
  FOT [P\textsubscript{99} = 30 ms] & \cellcolor{green} & \cellcolor{green} & \cellcolor{red} & \cellcolor{red} \\  
  \hline
  FOT [P\textsubscript{99} = 550 ms] & \cellcolor{red} & \cellcolor{red} & \cellcolor{red} & \cellcolor{red} \\
  \hline
  RRT* [P\textsubscript{99} = 15 ms] & \cellcolor{green} & \cellcolor{green} & \cellcolor{green} & \cellcolor{red} \\
  \hline
  RRT* [P\textsubscript{99} = 76 ms] & \cellcolor{red} & \cellcolor{red} & \cellcolor{red} & \cellcolor{red} \\
  \hline  
  Hybrid A* [P\textsubscript{99} = 25 ms] & \cellcolor{green} & \cellcolor{red} & \cellcolor{red} & \cellcolor{green} \\
  \hline  
  Hybrid A* [P\textsubscript{99} = 760 ms] & \cellcolor{red} & \cellcolor{red} & \cellcolor{red} & \cellcolor{red} \\  
\end{tabular}

  \label{t:planners}
  \vspace{-2em}
\end{table}

In this experiment, we omit the perception models and use ground truth to
compare each planner in isolation~\footnote{
  Visualizations are available at
  \url{https://tinyurl.com/y3coq57r}.\label{experiment-videos}
}.
For the Frenet Optimal Trajectory (FOT) planner~\cite{frenet-v1,frenet-v2}
we executed a fast and a slow configuration with $0.3$ seconds time
discretization and $0.5$ meters space discretization, respectively $0.1$ space
and time discretization.
Similarly, for RRT*~\cite{rrt-star} we experimented with $0.1$ and $0.5$ meters
step sizes
Lastly, we explore Hybrid A*~\cite{hybrid-astar} with step sizes of $3.0$
and $6.0$, and radian step discretizations of $0.25$ and $0.75$.

\begin{figure}
  \centering
  \includegraphics[scale=0.85]{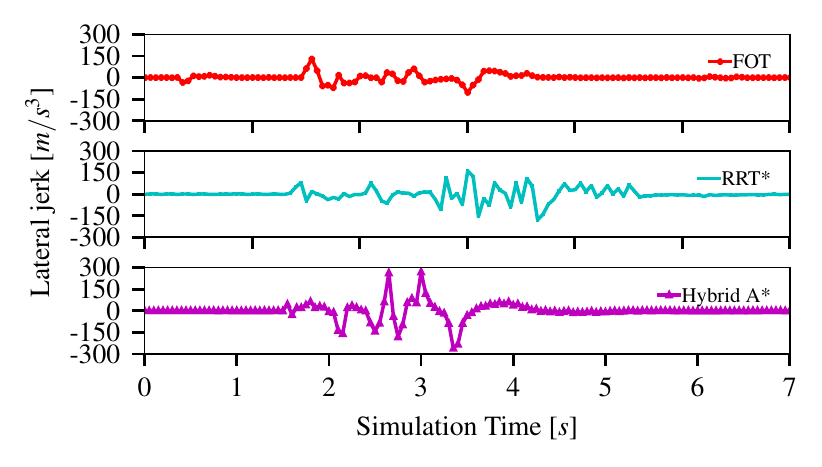}
  \caption{Comparison of the ride comfort offered by the planners that avoid
    the collision when driving at 16 m/s target speed.}
  \label{f:jerk-eval}
\end{figure}

In \cref{t:planners} we show which configurations avoided a collision with
the person and the truck for different \emph{target speeds} the \av{}
drove at before it detected the person.
Since avoiding a collision requires a fast, \emph{swerving} maneuver from the
\av{}, the planner configurations that minimized runtime performed better.
Furthermore, \cref{f:jerk-eval} compares the three configurations that succeed
at $16$m/s target speed by plotting the lateral jerk (i.e. ride comfort)
while performing the \emph{swerving} maneuver to avoid a collision.
We see that the three planners have markedly different jerk profiles.
While FOT avoids a collision in fewer cases than RRT*, it provides a more
comfortable ride.
This experiments illustrates the realistic end-to-end ``A/B testing'' of \av{}
components enabled by \pipeline{}.

\section{Conclusion}
\label{s:conclusions}
We have presented an open-source platform for \av{} research.
\pipeline{}'s modular and portable structure enables iterative development
and evaluation of components in the context of an entire \av{} pipeline.
This approach to \av{} development supports the study of interactions between
modules, resulting in a better understanding of the effects of runtime and
accuracy on end-to-end driving behavior.

\pipeline{} provides reference and ground-truth implementations for several
components of an \av{} pipeline.
These state-of-the-art implementations have been used to both drive a
real-vehicle and attain a high score on the \carla{} Challenge.

In the future, we aim to enable a similar study of reinforcement learning
based approaches to autonomous driving by providing an OpenAI Gym
interface~\cite{openai-gym}.
We hope that our work on \pipeline{} will help broader \av{} community
conduct impactful research that will lead to a safer autonomous future.

{

  \balance
  \bibliographystyle{IEEEtran}
  \bibliography{paper}

\begin{thebibliography}{10}
\providecommand{\url}[1]{#1}
\csname url@samestyle\endcsname
\providecommand{\newblock}{\relax}
\providecommand{\bibinfo}[2]{#2}
\providecommand{\BIBentrySTDinterwordspacing}{\spaceskip=0pt\relax}
\providecommand{\BIBentryALTinterwordstretchfactor}{4}
\providecommand{\BIBentryALTinterwordspacing}{\spaceskip=\fontdimen2\font plus
\BIBentryALTinterwordstretchfactor\fontdimen3\font minus
  \fontdimen4\font\relax}
\providecommand{\BIBforeignlanguage}[2]{{%
\expandafter\ifx\csname l@#1\endcsname\relax
\typeout{** WARNING: IEEEtran.bst: No hyphenation pattern has been}%
\typeout{** loaded for the language `#1'. Using the pattern for}%
\typeout{** the default language instead.}%
\else
\language=\csname l@#1\endcsname
\fi
#2}}
\providecommand{\BIBdecl}{\relax}
\BIBdecl

\bibitem{ren2015faster}
S.~Ren, K.~He, R.~Girshick, and J.~Sun, ``{Faster R-CNN}: Towards real-time
  object detection with region proposal networks,'' in \emph{Proceedings of the
  International Conferences on Advances in Neural Information Processing
  Systems (NeurIPS)}, 2015, pp. 91--99.

\bibitem{ssd}
W.~Liu, D.~Anguelov, D.~Erhan, C.~Szegedy, S.~Reed, C.-Y. Fu, and A.~C. Berg,
  ``{SSD}: Single shot multibox detector,'' in \emph{Proceedings of the
  European Conference on Computer Vision (ECCV)}.\hskip 1em plus 0.5em minus
  0.4em\relax Springer, 2016, pp. 21--37.

\bibitem{tan20efficientdet}
M.~Tan, R.~Pang, and Q.~V. Le, ``{EfficientDet}: Scalable and efficient object
  detection,'' in \emph{Proceedings of the IEEE Conference on Computer Vision
  and Pattern Recognition (CVPR)}, 2020.

\bibitem{rrt-star}
S.~Karaman and E.~Frazzoli, ``Sampling-based algorithms for optimal motion
  planning,'' \emph{The international journal of robotics research}, vol.~30,
  no.~7, pp. 846--894, 2011.

\bibitem{hybrid-astar}
D.~Dolgov, S.~Thrun, M.~Montemerlo, and J.~Diebel, ``Practical search
  techniques in path planning for autonomous driving,'' in \emph{Proceedings of
  the 1\textsuperscript{st} International Symposium on Search Techniques in
  Artificial Intelligence and Robotics (STAIR)}, vol. 1001, 2008, pp. 18--80.

\bibitem{frenet-v2}
W.~Xu, J.~Wei, J.~M. Dolan, H.~Zhao, and H.~Zha, ``A real-time motion planner
  with trajectory optimization for autonomous vehicles,'' in \emph{Proceedings
  of the IEEE International Conference on Robotics and Automation
  (ICRA)}.\hskip 1em plus 0.5em minus 0.4em\relax IEEE, 2012, pp. 2061--2067.

\bibitem{schwarting_planning_2018}
W.~Schwarting, J.~{Alonso-Mora}, and D.~Rus, ``\BIBforeignlanguage{en}{Planning
  and decision-making for autonomous vehicles},''
  \emph{\BIBforeignlanguage{en}{Annual Review of Control, Robotics, and
  Autonomous Systems}}, vol.~1, no.~1, pp. 187--210, May 2018.

\bibitem{guanetti_control_2018}
J.~Guanetti, Y.~Kim, and F.~Borrelli, ``\BIBforeignlanguage{en}{Control of
  connected and automated vehicles: {{State}} of the art and future
  challenges},'' \emph{\BIBforeignlanguage{en}{Annual Reviews in Control}},
  vol.~45, pp. 18--40, Jan. 2018.

\bibitem{lee2017desire}
N.~Lee, W.~Choi, P.~Vernaza, C.~B. Choy, P.~H. Torr, and M.~Chandraker,
  ``Desire: Distant future prediction in dynamic scenes with interacting
  agents,'' in \emph{Proceedings of the IEEE Conference on Computer Vision and
  Pattern Recognition (CVPR)}, 2017, pp. 336--345.

\bibitem{rhinehart2018r2p2}
N.~Rhinehart, K.~M. Kitani, and P.~Vernaza, ``{R2P2}: A reparameterized
  pushforward policy for diverse, precise generative path forecasting,'' in
  \emph{Proceedings of the European Conference on Computer Vision (ECCV)},
  2018, pp. 772--788.

\bibitem{mozaffari_deep_2020}
S.~Mozaffari, O.~Y. {Al-Jarrah}, M.~Dianati, P.~Jennings, and A.~Mouzakitis,
  ``Deep learning-based vehicle behavior prediction for autonomous driving
  applications: A review,'' \emph{IEEE Transactions on Intelligent
  Transportation Systems}, pp. 1--15, 2020.

\bibitem{janai_computer_2020}
J.~Janai, F.~G{\"u}ney, A.~Behl, and A.~Geiger,
  ``\BIBforeignlanguage{English}{Computer vision for autonomous vehicles:
  Problems, datasets and state of the art},''
  \emph{\BIBforeignlanguage{English}{Foundations and Trends in Computer
  Graphics and Vision}}, vol.~12, no. 1\textendash 3, pp. 1--308, Jul. 2020.

\bibitem{geiger_vision_2013}
A.~Geiger, P.~Lenz, C.~Stiller, and R.~Urtasun,
  ``\BIBforeignlanguage{en}{Vision meets robotics: {{The KITTI}} dataset},''
  \emph{\BIBforeignlanguage{en}{The International Journal of Robotics
  Research}}, vol.~32, no.~11, pp. 1231--1237, Sep. 2013.

\bibitem{cityscapes}
M.~Cordts, M.~Omran, S.~Ramos, T.~Rehfeld, M.~Enzweiler, R.~Benenson,
  U.~Franke, S.~Roth, and B.~Schiele, ``The {C}ityscapes dataset for semantic
  urban scene understanding,'' in \emph{Proceedings of the IEEE Conference on
  Computer Vision and Pattern Recognition (CVPR)}, 2016.

\bibitem{waymo-open-dataset}
\BIBentryALTinterwordspacing
P.~Sun, H.~Kretzschmar, X.~Dotiwalla, A.~Chouard, V.~Patnaik, P.~Tsui, J.~Guo,
  Y.~Zhou, Y.~Chai, B.~Caine, V.~Vasudevan, W.~Han, J.~Ngiam, H.~Zhao,
  A.~Timofeev, S.~Ettinger, M.~Krivokon, A.~Gao, A.~Joshi, S.~Zhao, S.~Cheng,
  Y.~Zhang, J.~Shlens, Z.~Chen, and D.~Anguelov, ``Scalability in perception
  for autonomous driving: Waymo open dataset,'' 2019. [Online]. Available:
  \url{https://arxiv.org/abs/1912.04838}
\BIBentrySTDinterwordspacing

\bibitem{argo-data}
``Argoverse,'' \url{https://www.argoverse.org/}.

\bibitem{cruise-report}
{General Motors}, ``{2018 Self-driving safety report},''
  \url{https://www.gm.com/content/dam/company/docs/us/en/gmcom/gmsafetyreport.pdf}.

\bibitem{ford-safety-report}
{Ford}, ``A matter of trust: {F}ord's approach to developing self-driving
  vehicles,''
  \url{https://media.ford.com/content/dam/fordmedia/pdf/Ford_AV_LLC_FINAL_HR_2.pdf}.

\bibitem{ntsb-uber}
``{NTSB's} accident report on the {U}ber self-driving vehicle crash.''
  \url{https://www.ntsb.gov/investigations/AccidentReports/Reports/HAR1903.pdf}.

\bibitem{autoware}
{Autoware}, ``{Autoware User's Manual - Document Version 1.1},''
  \url{https://github.com/CPFL/Autoware-Manuals/blob/master/en/Autoware_UsersManual_v1.1.md}.

\bibitem{apollo-baidu}
{Baidu}, ``{Apollo 3.0 Software Architecture},''
  \url{https://github.com/ApolloAuto/apollo/blob/master/docs/specs/Apollo_3.0_Software_Architecture.md}.

\bibitem{thrun2006stanley}
S.~Thrun, M.~Montemerlo, H.~Dahlkamp, D.~Stavens, A.~Aron, J.~Diebel, P.~Fong,
  J.~Gale, M.~Halpenny, G.~Hoffmann \emph{et~al.}, ``Stanley: The robot that
  won the {DARPA Grand Challenge},'' \emph{Journal of field Robotics}, vol.~23,
  no.~9, pp. 661--692, 2006.

\bibitem{philion2020learning}
J.~Philion, A.~Kar, and S.~Fidler, ``Learning to evaluate perception models
  using planner-centric metrics,'' in \emph{Proceedings of the IEEE Conference
  on Computer Vision and Pattern Recognition (CVPR)}, 2020, pp.
  14\,055--14\,064.

\bibitem{kitti-detection}
A.~Geiger, P.~Lenz, and R.~Urtasun, ``Are we ready for autonomous driving? the
  {KITTI} vision benchmark suite,'' in \emph{Proceedings of the IEEE Conference
  on Computer Vision and Pattern Recognition (CVPR)}, 2012.

\bibitem{streaming-perception}
M.~Li, Y.~Wang, and D.~Ramanan, ``Towards streaming perception,'' in
  \emph{Proceedings of the European Conference on Computer Vision (ECCV)}, Aug.
  2020.

\bibitem{erdos-code}
``{ERDOS}: Elastic robot data-flow operating system,''
  \url{https://github.com/erdos-project/erdos}.

\bibitem{carla-challenge}
``The {CARLA} autonomous driving challenge,''
  \url{https://leaderboard.carla.org/}.

\bibitem{speed-accuracy-detectors}
J.~Huang, V.~Rathod, C.~Sun, M.~Zhu, A.~Korattikara, A.~Fathi, I.~Fischer,
  Z.~Wojna, Y.~Song, S.~Guadarrama, and K.~Murphy, ``Speed/accuracy trade-offs
  for modern convolutional object detectors,'' in \emph{Proceedings of the IEEE
  Conference on Computer Vision and Pattern Recognition (CVPR)}, Jul. 2017.

\bibitem{yolov4}
\BIBentryALTinterwordspacing
A.~Bochkovskiy, C.-Y. Wang, and H.-Y. Mark~Liao, ``{YOLOv4}: Optimal speed and
  accuracy of object detection,'' 2020. [Online]. Available:
  \url{https://arxiv.org/abs/2004.10934}
\BIBentrySTDinterwordspacing

\bibitem{cruise-roscon}
N.~Valigi, ``Lessons learned building a self-driving car on ros,''
  \url{https://roscon.ros.org/2018/presentations/ROSCon2018_LessonsLearnedSelfDriving.pdf},
  2018.

\bibitem{krajzewicz2012recent}
D.~Krajzewicz, J.~Erdmann, M.~Behrisch, and L.~Bieker, ``Recent development and
  applications of sumo-simulation of urban mobility,'' \emph{International
  journal on advances in systems and measurements}, vol.~5, no. 3\&4, 2012.

\bibitem{carla}
A.~Dosovitskiy, G.~Ros, F.~Codevilla, A.~Lopez, and V.~Koltun, ``{CARLA}: {An}
  open urban driving simulator,'' in \emph{Proceedings of the
  1\textsuperscript{st} Conference on Robot Learning (CoRL)}, 2017, pp. 1--16.

\bibitem{airsim2017fsr}
\BIBentryALTinterwordspacing
S.~Shah, D.~Dey, C.~Lovett, and A.~Kapoor, ``{AirSim}: High-fidelity visual and
  physical simulation for autonomous vehicles,'' 2017. [Online]. Available:
  \url{https://arxiv.org/abs/1705.05065}
\BIBentrySTDinterwordspacing

\bibitem{altekar2009odr}
G.~Altekar and I.~Stoica, ``{ODR}: Output-deterministic replay for multicore
  debugging,'' in \emph{Proceedings of the 22\textsuperscript{nd} ACM Symposium
  on Operating Systems Principles (SOSP)}, 2009, pp. 193--206.

\bibitem{pytorch}
A.~Paszke, S.~Gross, F.~Massa, A.~Lerer, J.~Bradbury, G.~Chanan, T.~Killeen,
  Z.~Lin, N.~Gimelshein, L.~Antiga \emph{et~al.}, ``Pytorch: An imperative
  style, high-performance deep learning library,'' in \emph{Advances in Neural
  Information Processing Systems (NeurIPS)}, 2019, pp. 8026--8037.

\bibitem{tensorflow}
M.~Abadi, P.~Barham, J.~Chen, Z.~Chen, A.~Davis, J.~Dean, M.~Devin,
  S.~Ghemawat, G.~Irving, M.~Isard, M.~Kudlur, J.~Levenberg, R.~Monga,
  S.~Moore, D.~G. Murray, B.~Steiner, P.~Tucker, V.~Vasudevan, P.~Warden,
  M.~Wicke, Y.~Yu, and X.~Zheng, ``{TensorFlow}: A system for large-scale
  machine learning,'' in \emph{Proceedings of the 12\textsuperscript{th}
  USE\-NIX Symposium on Operating Systems Design and Implementation (OSDI)},
  Nov. 2016.

\bibitem{punctuations}
P.~A. Tucker, D.~Maier, T.~Sheard, and L.~Fegaras, ``Exploiting punctuation
  semantics in continuous data streams,'' \emph{IEEE Transactions on Knowledge
  and Data Engineering}, vol.~15, no.~3, pp. 555--568, 2003.

\bibitem{flink}
P.~Carbone, A.~Katsifodimos, S.~Ewen, V.~Markl, S.~Haridi, and K.~Tzoumas,
  ``Apache {Flink}: Stream and batch processing in a single engine,''
  \emph{Bulletin of the IEEE Computer Society Technical Committee on Data
  Engineering}, vol.~36, no.~4, 2015.

\bibitem{naiad}
D.~G. Murray, F.~McSherry, R.~Isaacs, M.~Isard, P.~Barham, and M.~Abadi,
  ``Naiad: A timely dataflow system,'' in \emph{Proceedings of the
  24\textsuperscript{th} ACM Symposium on Operating Systems Principles (SOSP)},
  Nov. 2013, pp. 439--455.

\bibitem{google-cloud-dataflow}
{Google Cloud Dataflow}. Google Inc. \url{http://cloud.google.com/dataflow/};
  accessed 11/11/2020.

\bibitem{google-sensor-throughput}
{Amara D. Angelica}, ``{Google’s self-driving car gathers nearly 1 GB/sec},''
  \url{http://www.kurzweilai.net/googles-self-driving-car-gathers-nearly-1-gbsec}.

\bibitem{rong2020lgsvl}
G.~Rong, B.~H. Shin, H.~Tabatabaee, Q.~Lu, S.~Lemke, M.~Mo{\v{z}}eiko,
  E.~Boise, G.~Uhm, M.~Gerow, S.~Mehta \emph{et~al.}, ``{LGSVL} simulator: A
  high fidelity simulator for autonomous driving,'' in \emph{Proceedings of the
  23\textsuperscript{rd} International Conference on Intelligent Transportation
  Systems (ITSC)}.\hskip 1em plus 0.5em minus 0.4em\relax IEEE, 2020, pp. 1--6.

\bibitem{carla-scenario}
``{ScenarioRunner for CARLA},''
  \url{https://github.com/carla-simulator/scenario_runner}.

\bibitem{nhtsa-scenarios}
{National Highway Traffic Safety Administration}, ``{Pre-Crash Scenario
  Typology for Crash Avoidance Research},''
  \url{https://www.nhtsa.gov/sites/nhtsa.dot.gov/files/pre-crash_scenario_typology-final_pdf_version_5-2-07.pdf}.

\bibitem{google-dataflow}
\BIBentryALTinterwordspacing
T.~Akidau, R.~Bradshaw, C.~Chambers, S.~Chernyak, R.~J.
  Fern\'{a}ndez-Moctezuma, R.~Lax, S.~McVeety, D.~Mills, F.~Perry, E.~Schmidt,
  and S.~Whittle, ``The dataflow model: A practical approach to balancing
  correctness, latency, and cost in massive-scale, unbounded, out-of-order data
  processing,'' \emph{Proceedings of the VLDB Endowment}, vol.~8, no.~12, pp.
  1792--1803, Aug. 2015. [Online]. Available:
  \url{http://dx.doi.org/10.14778/2824032.2824076}
\BIBentrySTDinterwordspacing

\bibitem{kahn-process-network}
K.~Gilles, ``The semantics of a simple language for parallel programming,''
  \emph{Information processing}, vol.~74, pp. 471--475, 1974.

\bibitem{chrome-profiling}
``The trace event profiling tool (about:tracing),''
  \url{https://www.chromium.org/developers/how-tos/trace-event-profiling-tool}.

\bibitem{philion2020lift}
J.~Philion and S.~Fidler, ``Lift, splat, shoot: Encoding images from arbitrary
  camera rigs by implicitly unprojecting to {3D},'' in \emph{Proceedings of the
  European Conference on Computer Vision (ECCV)}.\hskip 1em plus 0.5em minus
  0.4em\relax Springer, 2020, pp. 194--210.

\bibitem{tracking_sort}
A.~Bewley, Z.~Ge, L.~Ott, F.~Ramos, and B.~Upcroft, ``Simple online and
  realtime tracking,'' in \emph{Proceedings of the 23\textsuperscript{th} IEEE
  International Conference on Image Processing (ICIP)}, 2016, pp. 3464--3468.

\bibitem{tracking_deep_sort}
N.~Wojke, A.~Bewley, and D.~Paulus, ``Simple online and realtime tracking with
  a deep association metric,'' in \emph{Proceedings of the
  24\textsuperscript{th} IEEE International Conference on Image Processing
  (ICIP)}.\hskip 1em plus 0.5em minus 0.4em\relax IEEE, 2017, pp. 3645--3649.

\bibitem{tracking_da_siam_rpn}
Z.~Zhu, Q.~Wang, L.~Bo, W.~Wu, J.~Yan, and W.~Hu, ``Distractor-aware siamese
  networks for visual object tracking,'' in \emph{Proceedings of the European
  Conference on Computer Vision (ECCV)}, 2018.

\bibitem{chai2019multipath}
\BIBentryALTinterwordspacing
Y.~Chai, B.~Sapp, M.~Bansal, and D.~Anguelov, ``{MultiPath}: Multiple
  probabilistic anchor trajectory hypotheses for behavior prediction,'' 2019.
  [Online]. Available: \url{https://arxiv.org/abs/1910.05449}
\BIBentrySTDinterwordspacing

\bibitem{tang2019multiple}
C.~Tang and R.~R. Salakhutdinov, ``Multiple futures prediction,'' in
  \emph{Proceedings of the International Conference on Advances in Neural
  Information Processing Systems (NeurIPS)}, 2019, pp. 15\,398--15\,408.

\bibitem{lavalle-planning}
S.~M. LaValle, \emph{Planning Algorithms}.\hskip 1em plus 0.5em minus
  0.4em\relax USA: Cambridge University Press, 2006.

\bibitem{katrakazas}
C.~Katrakazas, M.~Quddus, W.-H. Chen, and L.~Deka, ``Real-time motion planning
  methods for autonomous on-road driving: State-of-the-art and future research
  directions,'' \emph{Transportation Research Part C: Emerging Technologies},
  vol.~60, pp. 416--442, 2015.

\bibitem{paden-survey}
B.~Paden, M.~{\v{C}}{\'a}p, S.~Z. Yong, D.~Yershov, and E.~Frazzoli, ``A survey
  of motion planning and control techniques for self-driving urban vehicles,''
  \emph{IEEE Transactions on intelligent vehicles}, vol.~1, no.~1, pp. 33--55,
  2016.

\bibitem{frenet-v1}
M.~Werling, J.~Ziegler, S.~Kammel, and S.~Thrun, ``Optimal trajectory
  generation for dynamic street scenarios in a frenet frame,'' in
  \emph{Proceedings of the IEEE International Conference on Robotics and
  Automation (ICRA)}.\hskip 1em plus 0.5em minus 0.4em\relax IEEE, 2010, pp.
  987--993.

\bibitem{amos2018differentiable}
B.~Amos, I.~D.~J. Rodriguez, J.~Sacks, B.~Boots, and J.~Z. Kolter,
  ``Differentiable mpc for end-to-end planning and control,'' in
  \emph{Proceedings of the 32\textsuperscript{nd} International Conference on
  Neural Information Processing Systems (NeurIPS)}, 2018, pp. 8299--8310.

\bibitem{liu2017path}
C.~Liu, S.~Lee, S.~Varnhagen, and H.~E. Tseng, ``Path planning for autonomous
  vehicles using model predictive control,'' in \emph{Proceedings of the IEEE
  Intelligent Vehicles Symposium (IV)}.\hskip 1em plus 0.5em minus 0.4em\relax
  IEEE, 2017, pp. 174--179.

\bibitem{liang2020pnpnet}
M.~Liang, B.~Yang, W.~Zeng, Y.~Chen, R.~Hu, S.~Casas, and R.~Urtasun,
  ``{PnPNet}: End-to-end perception and prediction with tracking in the loop,''
  in \emph{Proceedings of the IEEE/CVF Conference on Computer Vision and
  Pattern Recognition (CVPR)}, 2020, pp. 11\,553--11\,562.

\bibitem{zeng2019end}
W.~Zeng, W.~Luo, S.~Suo, A.~Sadat, B.~Yang, S.~Casas, and R.~Urtasun,
  ``End-to-end interpretable neural motion planner,'' in \emph{Proceedings of
  the IEEE/CVF Conference on Computer Vision and Pattern Recognition (CVPR)},
  2019, pp. 8660--8669.

\bibitem{wayve-website}
``Wayve.ai,'' \url{https://wayve.ai}.

\bibitem{dataspeed}
``Dataspeed: The industry-leading drive-by-wire kit,''
  \url{https://www.dataspeedinc.com/}.

\bibitem{biber2003normal}
P.~Biber and W.~Stra{\ss}er, ``The normal distributions transform: A new
  approach to laser scan matching,'' in \emph{Proceedings of the IEEE/RSJ
  International Conference on Intelligent Robots and Systems (IROS)},
  vol.~3.\hskip 1em plus 0.5em minus 0.4em\relax IEEE, 2003, pp. 2743--2748.

\bibitem{kitti-website}
A.~Geiger, P.~Lenz, C.~Stiller, and R.~Urtasun, ``{The KITTI Vision Benchmark
  Suite},'' \url{http://www.cvlibs.net/datasets/kitti/}.

\bibitem{openai-gym}
\BIBentryALTinterwordspacing
G.~Brockman, V.~Cheung, L.~Pettersson, J.~Schneider, J.~Schulman, J.~Tang, and
  W.~Zaremba, ``{OpenAI gym},'' 2016. [Online]. Available:
  \url{https://arxiv.org/abs/1606.01540}
\BIBentrySTDinterwordspacing

\end{thebibliography}
}

\end{document}